\documentclass[lettersize,journal]{IEEEtran}
\usepackage{amsmath,amsfonts}
\usepackage{algorithmic}
\usepackage{algorithm}
\usepackage{array}
\usepackage[caption=false,font=normalsize,labelfont=sf,textfont=sf]{subfig}
\usepackage{textcomp}
\usepackage{stfloats}
\usepackage{url}
\usepackage{verbatim}
\usepackage{graphicx}
\usepackage{booktabs}
\usepackage{cite}

\begin{document}

\title{PRG: Prompt-Based Distillation Without Annotation via Proxy Relational Graph}

\author{Yijin Xu, Jialun Liu, Hualiang Wei, and Wenhui Li
\thanks{Yijin Xu, Hualiang Wei, and Wenhui Li are with the College of Computer Science and Technology, and the Key Laboratory of Symbolic Computation and Knowledge Engineer, Ministry of Education, Jilin University, Changchun, Jilin 130012, China (e-mail: xuyj21@mails.jlu.edu.cn;
liwh@mails.jlu.edu.cn).}
\thanks{Jialun Liu is with Baidu VIS., Beijing 100000, China (e-mail:
liujialun@baidu.com)}
}

\markboth{Journal of \LaTeX\ Class Files,~Vol.~14, No.~8, August~2021}%
{Shell \MakeLowercase{\textit{et al.}}: A Sample Article Using IEEEtran.cls for IEEE Journals}

\IEEEpubid{0000--0000/00\$00.00~\copyright~2021 IEEE}

\maketitle

\begin{abstract}
In this paper, we propose a new distillation method for extracting knowledge from Large Foundation Models (LFM) into lightweight models, introducing a novel supervision mode that does not require manually annotated data. While LFMs exhibit exceptional zero-shot classification abilities across datasets, relying solely on LFM-generated embeddings for distillation poses two main challenges: LFM's task-irrelevant knowledge and the high density of features. The transfer of task-irrelevant knowledge could compromise the student model's discriminative capabilities, and the high density of features within target domains obstructs the extraction of discriminative knowledge essential for the task.
To address this issue, we introduce the Proxy Relational Graph (PRG) method. We initially extract task-relevant knowledge from LFMs by calculating a weighted average of logits obtained through text prompt embeddings. Then we construct sample-class proxy graphs for LFM and student models, respectively, to model the correlation between samples and class proxies. Then, we achieve the distillation of selective knowledge by aligning the relational graphs produced by both the LFM and the student model. Specifically, the distillation from LFM to the student model is achieved through two types of alignment: 1) aligning the sample nodes produced by the student model with those produced by the LFM, and 2) aligning the edge relationships in the student model's graph with those in the LFM's graph. 
Our experimental results validate the effectiveness of PRG, demonstrating its ability to leverage the extensive knowledge base of LFMs while skillfully circumventing their inherent limitations in focused learning scenarios. Notably, in our annotation-free framework, PRG achieves an accuracy of 76.23\% (T: 77.9\%) on CIFAR-100 and 72.44\% (T: 75.3\%) on the ImageNet-1K.
\end{abstract}


\begin{figure}[ht]
  \centering
  \includegraphics[width=\linewidth]{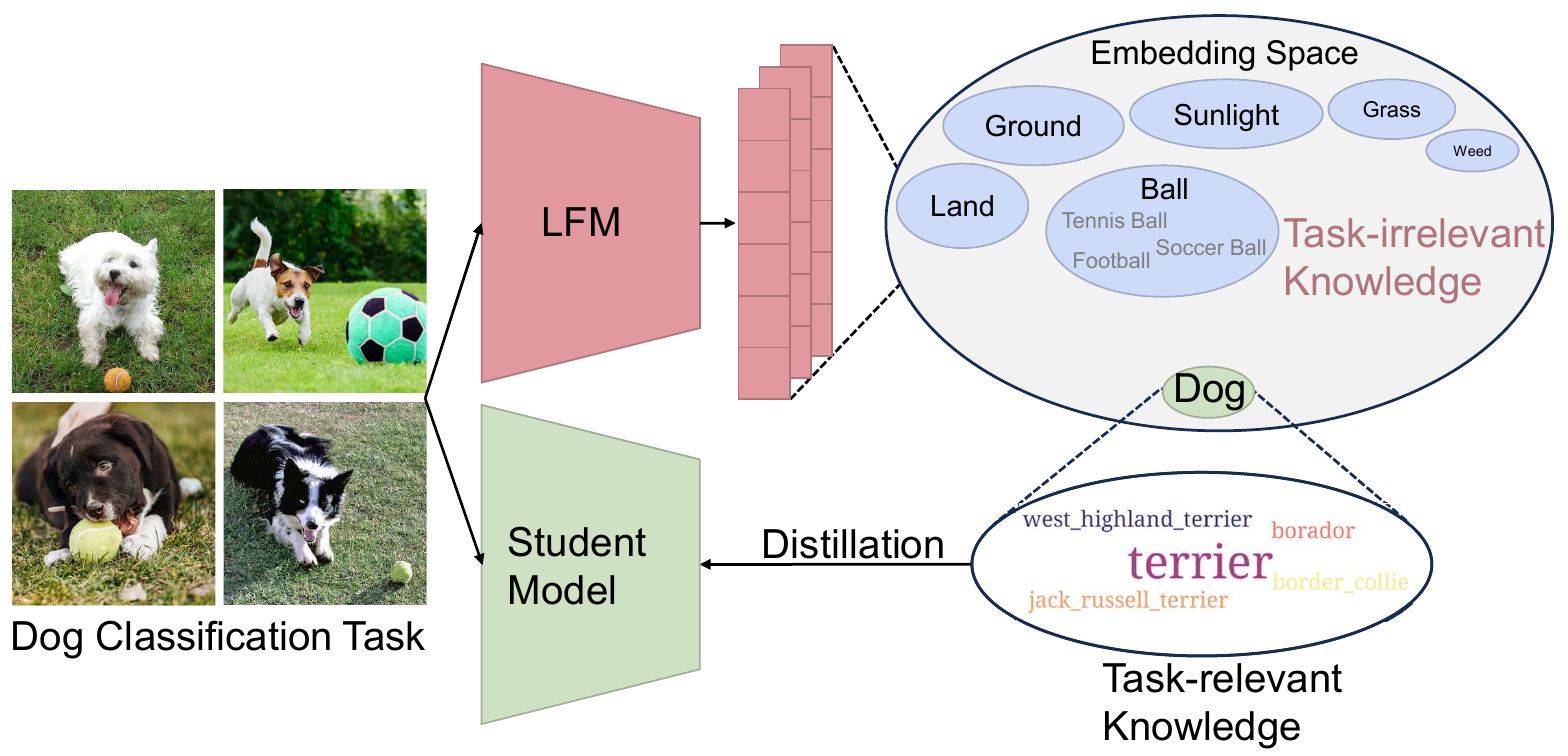}
  \caption{Using a Large Foundation Model (LFM) as a teacher for knowledge distillation in a dog classification task introduces challenges: 1) LFM captures irrelevant information like grass and balls, which are not needed for fine-grain classification and should not be distilled. 2) The LFM's vast feature space means that dog-specific features occupy a small fraction, making it hard for the student to learn discriminative knowledge in such a dense space.}
  \label{fig:banner}
\end{figure}

\section{Introduction}
\IEEEPARstart{K}{nowledge}  Distillation (KD) has emerged as a potent tool for compressing models and transferring knowledge, showcasing significant accomplishments across various downstream tasks including image classification, object detection, and instance segmentation. 

Traditional knowledge distillation methodologies have two primary constraints: 
\textbf{Data Annotation} requires comprehensive labeling of samples within the target domain. \textbf{Teacher Pre-training} involves the pre-training of large teacher models on datasets specific to the target domain.
These constraints highlight the need for universally deployable and highly adaptable KD methodologies, enabling efficient utilization and implementation across a broader spectrum of tasks. Consequently, this opens the door to leveraging the unique capabilities of large foundational models (LFMs). Foundational models, such as CLIP \cite{radford2021learning}, demonstrate their capacity to assimilate knowledge from extensive volumes of data. The broad and discriminatively rich feature spaces developed by these models facilitate the encapsulation of a wide spectrum of general knowledge. Their cross-task capability unveils novel pathways for knowledge distillation, eliminating the necessity for specialized pre-trained teacher models in specific domains. Additionally, the advanced cognitive abilities of LFMs across various domains support the distillation of knowledge from unannotated data.

\IEEEpubidadjcol
Using LFMs as teacher models and guiding student models solely with the teacher's zero-shot outputs on target datasets is a potentially effective distillation strategy. This approach essentially facilitates unsupervised learning on the target dataset in a cost-effective and efficient manner. However, distilling the comprehensive knowledge inherent in LFMs into smaller, domain-specific student models without the aid of manual annotations introduces a series of distinct challenges: 
1) \textbf{Task-irrelevant Knowledge}: A major challenge is separating useful domain-specific knowledge from the vast general knowledge within LFMs. These models contain extensive information, much of which is irrelevant for the specific task, risking the overload of the student model, thereby straining its limited capacity. As illustrated in Figure~\ref{fig:banner}, when using an LFM for distillation on a fine-grained dog dataset, the embeddings generated might include irrelevant background elements like grass and balls, which should not be transferred to the student.
2) \textbf{High Density of Features}: LFMs, trained on broad datasets, develop densely packed feature spaces. This density complicates feature-level distillation, particularly in fine-grained classifications where it can obscure essential subtle differences, thus impairing distillation effectiveness. As shown in Figure~\ref{fig:banner}, in such scenarios, dog-related features might occupy only a small portion of the LFM's feature space, making it difficult for the student to learn discriminative knowledge about dogs from such a congested feature space.

Specifically, our approach begin with an annotation-free knowledge distillation framework, utilizing the CLIP model as the teacher. Within this framework, the logits from CLIP on the target dataset serve as the supervisory signal for regular classification loss and as soft labels for KD loss. We assess the effectiveness of the conventional KD method within this framework, revealing a notable gap in accuracies between the student and CLIP. This indicates an incomplete acquisition of the discriminative knowledge embedded in CLIP's target task.

Simultaneously, it is observed that the choice of text prompts in the CLIP teacher model significantly influenced the outcomes of knowledge distillation. Prompts that provide more precise descriptions of images in the dataset yield superior distillation effects. This critical observation lead us to infer that CLIP contains both task-relevant and task-irrelevant knowledge with respect to the target dataset for distillation. More accurate prompts facilitate the representation of a greater amount of task-relevant knowledge within CLIP's zero-shot logits. It is the distillation of this task-relevant knowledge that proves beneficial for student models addressing target tasks, thereby catalyzing the research presented in this paper.

Building on this, we introduce the Proxy Relational Graph (PRG) method to refine the focus of distillation on task-relevant knowledge. In our approach, sample nodes capture the knowledge of individual samples by combining image features and logits outputs, addressing the issue of high feature density often seen in LFMs within target domains. However, directly aligning the student’s sample nodes with those of the LFM can inadvertently transfer task-irrelevant knowledge. To address this, we assign weights to the zero-shot logits based on text prompt embeddings, considering their classification performance on samples. This utilizes CLIP’s textual modality to extract and encapsulate task-relevant knowledge within the sample nodes. Furthermore, class proxy nodes represent the category-specific knowledge of the target task by continuously updating these nodes based on sample nodes that CLIP identifies as belonging to the corresponding category. As proxy nodes are not directly generated by the student model, aligning class proxy nodes is impractical. Therefore, we connect sample nodes and class proxy nodes with edges by calculating the correlation matrix between the sample nodes and class proxy nodes. These edges depict the sample-class relationships. 
By aligning these edges, we not only effectively filter out task-irrelevant knowledge that might be present in conventional relational graphs, but also transfer task-relevant relational knowledge to the student model.

In our experimental evaluation, we extensively test the proposed PRG method on image classification tasks to confirm its superiority. Across various networks and task datasets, PRG consistently achieves significant accuracy boosts, outperforming other existing methods by large margins. For example, on the CIFAR-100 dataset, PRG shows accuracy gains ranging from 1.21\% to 1.42\% compared to KD \cite{hinton2015distilling}. Additionally, experiments on the ImageNet-1K\cite{deng2009imagenet} dataset reveal a 0.78\% accuracy enhancement with PRG compared to KD, highlighting the broad applicability and effectiveness of our approach.

In summary, our work makes several significant contributions to the field of knowledge distillation:
\begin{itemize}
\item We are the first to employ foundational models as teachers for direct knowledge distillation into student models across various target image classification tasks without annotations. Additionally, we have devised a framework to distill the CLIP model across different datasets without requiring annotations.

\item We have recognized that the primary challenge of using LFMs for knowledge distillation in specific target tasks without annotations revolves around accurately modeling and transferring task-relevant knowledge. In response, we have developed the Proxy Relational Graph (PRG) method, which establishes relationships between samples and class proxies to integrate relational task-relevant knowledge. Moreover, we leverage the textual modality of CLIP to inject individual task-relevant knowledge into the PRG nodes.

\item The PRG method demonstrates superior performance across multiple configurations and datasets, encompassing both standard and fine-grained classification tasks. Particularly, our experiments on the ImageNet-1K\cite{deng2009imagenet}  dataset validate the effectiveness of PRG in enhancing the annotation-free distillation process, showcasing its broad applicability and robustness.

\end{itemize}

\begin{figure*}[!t]
\centering
\includegraphics[width=0.95\textwidth]{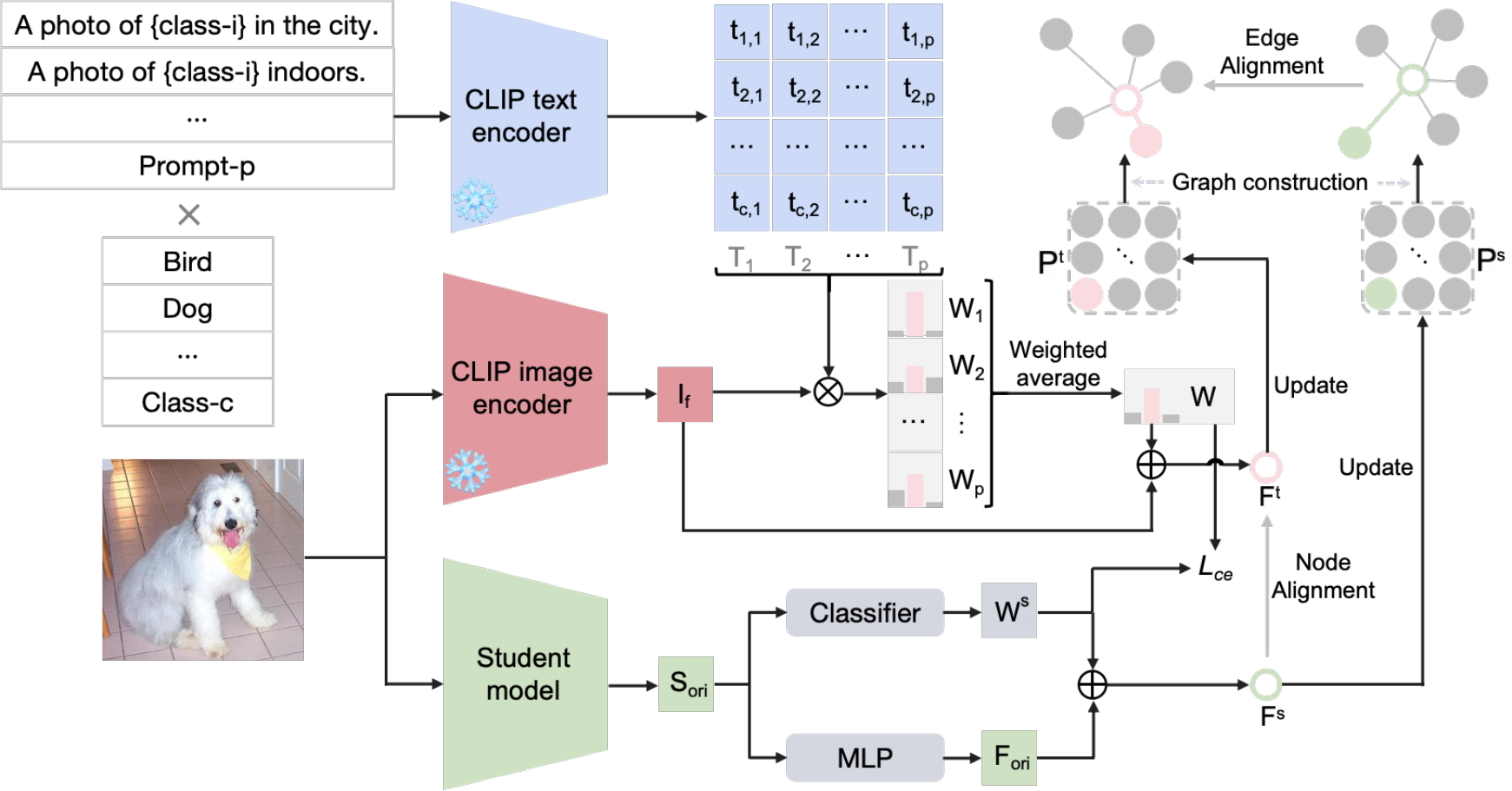}
\caption{Overview of our method: Text prompts are encoded via the CLIP text encoder to generate classification weights \(T\), and sample images are processed via the CLIP image encoder to obtain teacher image features \(I_f\). Each \(T_i\) multiplies \(I_f\) to produce logits \(W_i\) for corresponding prompts. \(W_i\) are weighted by their max logit and averaged to create weighted logits \(W\), which concatenated with \(I_f\) to form the teacher's sample node \(F^t\).
Simultaneously, the student model processes images to produce features \(F_{\text{ori}}\) that match the dimensions of \(I_f\) and classification logits \(W^s\), which are then concatenated to form the student sample node \(F^s\). Class proxy nodes \(P^t\) and \(P^s\) are updated by sample nodes. Pearson correlation coefficients between sample and class proxy nodes form relational graph edges \(E^t\) and \(E^s\). Finally, node and edge alignments transfer task-relevant knowledge to student model, optimizing distillation effectiveness.}
\label{fig:main}
\end{figure*}

\section{Related Works}
Knowledge distillation (KD) is a common approach to model compression, which involves training a student model using knowledge from a high-capacity teacher model to achieve better performance and accuracy.This is a process of knowledge transfer, transferring from the teacher network to the student network, training the student model to mimic the behavior and predictions of the teacher model by transferring knowledge.Based on the different transfer methods,it can be broadly classified into three main directions: logit-based, feature-based, and relation-based. Early works like \cite{hinton2015distilling,zhang2018deep,cho2019efficacy,yang2019training,phuong2019distillation,buciluǎ2006model} used soft logit outputs from teachers for additional supervision. Subsequent feature distillation methods \cite{adriana2015fitnets,zagoruyko2017paying,yim2017gift,heo2019comprehensive,chung2020feature,shu2021channel} focused on intermediate feature representations. Relation distillation methods\cite{park2019relational,tian2019crd,tung2019similarity,liu2019knowledge} explored the structural information in teachers’ logits or features. Other approaches\cite{zhang2019your,zhang2021self,bai2020few,liu2020adaptive,yang2021adversarial} have investigated different training paradigms and teacher model generation strategies. Among them, the first knowledge distillation method \cite{hinton2015distilling}employs softened labels to learn the class distribution. It proposes a variation of softmax and introduces a variable T to generate softened labels.DML\cite{zhang2018deep}breaks the predefined "strong-weak relationship" of traditional distillation models, enabling a group of student networks to learn from and guide each other during training.\cite{cho2019efficacy}propose an early-stop teacher regularization method for distillation, which involves stopping the distillation process early when approaching convergence.In\cite{yang2019training}, KD softened labels are further used, and constraints are added during the distillation process to optimize the objective.Due to the poor performance when there is a large capacity gap between the student and teacher network models,\cite{phuong2019distillation}borrow from muti-exit architectures to perform ensemble distribution knowledge distillation, extending to propose a new loss function. TinyViT \cite{wu2022tinyvit} distills small student models by pre-sparsifying and storing the logits of large teacher models, thereby saving memory and computational overhead.

Feature distillation methods align student feature maps with those of the teacher across varying dimensions. Fitnets \cite{adriana2015fitnets} pioneered this approach by using intermediate feature layers for knowledge transfer. To enhance generalization, \cite{zagoruyko2017paying} use attention maps from the teacher to guide the student, increasing the adaptability of the student's features. Additionally, \cite{chung2020feature} introduces adversarial perturbations during training to boost the student's capability to extract knowledge from the teacher.

Relation-based KD views knowledge as a graph with feature embeddings as nodes. Methods like DarkRank \cite{chen2018darkrank}, MHGD \cite{lee2019graph}, RKD \cite{park2019relational}, and CCKD \cite{peng2019correlation} employ various strategies to generate edge weights, primarily using similarity metrics. Contrastive and self-supervised learning approaches such as CRD \cite{tian2019crd}, CRCD \cite{zhu2021complementary}, SSKD \cite{xu2020knowledge}, and CSKD \cite{chen2020improving} aim to align representations and integrate category-level information. Techniques like LKD \cite{li2020local} and GLD \cite{kim2021distilling} emphasize local features and relationships, showcasing the range of distillation strategies.

In contrast to these methods, our approach focuses on distilling with the foundation models without manually annotated data and exploits its potential.

\section{Method}
\subsection{Overview}
This paper introduces the Proxy Relational Graph (PRG), a novel relational knowledge structure designed to facilitate the efficient transfer of task-relevant knowledge from LFMs to student models without annotation. To this end, we present a weakly supervised distillation framework for training student models on any dataset, thereby eliminating the need for artificial annotations through the use of LFMs as a teacher model. As illustrated in Figure~\ref{fig:main}, the process begins with the assembly of class names and prompt templates pertinent to the task into class text prompts. Subsequently, the prompts are then processed through the CLIP text encoder to generate text embedding matrices \(T\). Each matrix \(T_i\) serves as the weight vector for the \(i\)-th classifier , which is associated with its corresponding prompt. These weights are then multiplied by the image features \(I_f\) produced by CLIP, resulting in the generation of \(p\) sets of logits \(W_i\). Each set \(W_i\) represents the classification predictions based on the \(i\)-th prompt. It resides in \(\mathbb{R}^{c \times d}\), with \(c\) denoting the number of classification categories and \(d\) representing the dimension of the features output by CLIP. The maximum logit value from each \(W_i\) is employed as a weight to compute a weighted average, forming \(W\), which refines task-relevant knowledge by integrating semantic information from CLIP's text modality. The softmax-normalised version of \(W\) acts as a supervisory signal, used to compute the cross-entropy loss \(L_{ce}\) alongside the student classifier’s logits \(W_s\).

To construct the PRG nodes, we fuse \(I_f\) with \(W\) to create the teacher’s sample node \(F^t\). Similarly, the student's original image features \(S_{\text{ori}}\) are transformed via an MLP to \(F_{\text{ori}}\), ensuring dimensional alignment with \(I_f\). Then, \(F_{\text{ori}}\) is fused with \(W^s\) to form \(F^s\). Node Alignment involves aligning \(F^s\) with \(F^t\) to transfer encapsulated task-relevant knowledge to the student.

Additionally, PRG initializes class proxy nodes for each category as \(P^t\) and \(P^s\) for the teacher and student models, respectively. The classification of a sample into category \(j\) by CLIP prompts slight updates to \(P_j^t\) by \(F^t\) and to \(P_j^s\) by \(F^s\). This updating mechanism filters out task-irrelevant knowledge while retaining essential category-specific insights. Edges, quantified by Pearson correlation coefficients (PCC) \cite{pearson1896mathematical} between sample nodes and class proxy nodes, form the sets \(E^t(F^t,P^t)\) and \(E^s(F^s,P^s)\). Edge Alignment then aligns \(E^s(F^s,P^s)\) with \(E^t(F^t,P^t)\), facilitating the transfer of task-relevant relational knowledge. This overview succinctly summarizes the PRG distillation process, illustrating its efficiency in knowledge transfer.

\subsection{Framework for Annotation-free Knowledge Distillation}
The foundation of all methods in this work is a framework for knowledge distillation training that does not utilize annotations. This framework primarily relies on a Large Foundation Model (LFM) teacher to provide the supervisory signal. In this case, CLIP is chosen as the LFM teacher model due to its powerful zero-shot classification capability, enabling it to encode and classify images from any dataset. Its combination of visual and textual modalities allows it to provide rich and useful supervision signals. Through CLIP's text modality and zero-shot classification process, classification logits \(W\) are output for input images on any dataset. Soft labels for the given sample are obtained by applying the softmax function to these logits. Soft labels provide richer knowledge compared to hard labels by including probability distribution information, which helps the student model learn more about class relationships. They also mitigate the impact of noise from CLIP's incorrect classifications, enhancing the robustness of the learning process. These soft labels supervise the student model, and the student model outputs logits \(W^s\). The soft cross-entropy loss \(L_{\text{ce}}\) is calculated as follows:

\begin{equation}
L_{\text{ce}} = -\frac{1}{b} \sum_{i=1}^{b} \sum_{j=1}^{c} W_{ij} \log \left(\frac{\exp(W^s_{ij})}{\sum_{k=1}^{c} \exp(W^s_{ik})}\right)
\end{equation}
Where \(b\) is the batch size and \(c\) is the number of classes, \(W_{ij}\) represents the value of the softmax-processed logits from the teacher model for the \(i\)-th sample's \(j\)-th class. Meanwhile, \(W^s_{ij}\) denotes the logits predicted by the student model for the \(i\)-th sample as the \(j\)-th class. Thus, a preliminary framework for annotation-free knowledge distillation is established. On this basis, existing knowledge distillation methods can further distill knowledge by leveraging the image features or classification logits, or a combination of both, provided by the CLIP model. The overall loss function is expressed as: 
\begin{equation}
L = L_{\text{ce}} + L_{\text{kd}}
\end{equation}
where the calculation of \(L_{\text{kd}}\) depends on the specific method used.

\subsection{Construction of PRG}
\label{sec:graph}

In our Proxy Relational Graph (PRG) approach, we first modify the zero-shot classifier of CLIP to get a weighted \(W\) based on prompts. Then we concatenate each sample's logits and features to form integrated embeddings as sample nodes \(F\). Since logits are based on the target task, it enriches the task-related knowledge in the integrated embeddings. Then, we construct a graph \(\mathcal{G} = (N, E)\) to capture the complex structural information within the integrated embedding space. The node set \(N\) comprises two subsets: sample nodes \(F\) and class proxy nodes \(P\), where \(N = F \cup P\). The edge set \(E\) establishes structural correlations among these nodes, defined as \(E\subseteq {(\mathbf{f}_i,\mathbf{p}_j)\mid \mathbf{f}_i\in F, \mathbf{p}_j\in P}\). The details of the graph construction are as follows. 

\noindent\textbf{Weighted Logits Based on Prompts.} Initially, for the output of the CLIP model, we construct prompt-based weighted logits by modifying the zero-shot classifier structure of CLIP. Firstly, in selecting prompt templates for specific datasets, we adhere to the templates provided by CLIP for those particular datasets. Assuming the number of categories is \(c\), the number of prompt templates is \(p\), and the embedding dimension of CLIP is \(d\), then the assembled text prompts, processed through CLIP's Text Encoder, yield text embeddings \(T \in \mathbb{R}^{c \times p \times d}\). The original CLIP classifier averages the embeddings of all prompts by category to obtain \(T_{\text{cls}} \in \mathbb{R}^{c \times d}\) as the weights for the zero-shot classifier. Instead, we treat \(T\) as the weights of \(p\) classifiers:
\begin{equation*}
T = \{T_1, T_2, \ldots, T_p\}, T_{i} \in \mathbb{R}^{c \times d}
\end{equation*}
Considering a specific sample, we compute the logits by multiplying the output \(I_f\) from the CLIP Image Encoder with \(T\), resulting in \(p\) logits, \(W_1, W_2, \ldots, W_p\).
The logits generated by each prompt vary with the accuracy of the prompt's description of the image. We postulate that the larger the value of \(\max(W_i)\) among all logits, the more accurate the prompt's description of the sample, and thus, the richer the task-relevant knowledge contained in \(W_i\). Therefore, the final \(W\) used for node construction is formed as follows:
\begin{equation}
W = \sum_{i=1}^{p} w_i \cdot W_i
\label{eq:lt}
\end{equation}
where
\begin{equation}
w_i = \frac{\max(W_i)}{\sum_{j=1}^{p} \max(W_j)}
\end{equation}
This approach allows \(W\), compared to directly using CLIP's zero-shot logits, to acquire more abundant task-relevant knowledge from CLIP's textual modality through text prompts, thereby enhancing the distillation effect. Concurrently, the student model's output logits, denoted as \(W^s\), are produced through its classifier without any modifications, maintaining a standard process for the student's logits generation.

Subsequently, the original image features \(S^{\text{ori}}\) from the student model are aligned with the teacher's features \(I_f\) using a MLP, resulting in the transformed student features \(F_{ori}\), which reside in a \(d\)-dimensional space. This step ensures that the dimensions are comparable.

\noindent\textbf{Node Construction.} Upon completing the preparatory steps, we acquire the logits and features necessary for constructing nodes. For each model, we concatenate the logits and features to form integrated embeddings, which then act as nodes for individual samples within their respective models. Each of these integrated embeddings resides in \(\mathbb{R}^{D}\), where \(D = d + c\).
 The set of sample nodes \(F\) can then be defined as:
\begin{equation*}
F = \{ \mathbf{f}_{1}, \mathbf{f}_{2}, ..., \mathbf{f}_{b}\}
\end{equation*}
where \(b\) is the batch size in the distillation training process. Consequently, this process establishes the instance node sets \(F^s\) and \(F^t\) for the student and teacher models, respectively.
Simultaneously, we construct the class proxy nodes \(c\), each residing in \(\mathbb{R}^{D}\), with the set of all class proxy nodes denoted as \(P\):

\begin{equation*}
    P = \{\mathbf{p}_1, \mathbf{p}_2, \ldots, \mathbf{p}_c\}
\end{equation*}
This formulation applies consistently to both the teacher and student class proxy nodes, denoted as \(P^t\) and \(P^s\), respectively. The construction of these class proxy embeddings employs a progressive update approach, dynamically adapting the randomly initialized class proxies based on the integrated embeddings for each batch. Further details are provided subsequently.

\noindent\textbf{Configuration of Class Proxy Nodes.} As outlined in the beginning of Section~\ref{sec:graph}, the node set \(N\) is defined as \(N = F \cup P\), where \(F\) represents the node embeddings for a batch, and \(P\) represents the class proxy nodes. The specific methodology for constructing the class proxy embedding nodes \(P\) is detailed below. Before the start of the training phase, the class proxy embedding matrices for both the teacher and student networks, denoted as \(P^t\) and \(P^s\), are initialized randomly following a standard normal distribution. These matrices have dimensions \(c \times D\).

Let \(P^t_{(i, k)}\) and \(P^s_{(i, k)}\) represent the class proxy nodes for the \(i\)-th category in the teacher and student networks, respectively, at the \(k\)-th iteration. At each training iteration, these nodes are updated based on the sample node embeddings of the current batch. The update rule can be formalized as follows:

\begin{equation}
\label{eq:proxy_update}
P_{(i, k)} = P_{(i, k-1)} + \alpha \cdot \Delta P_{(i, k-1)}
\end{equation}

where \(\Delta P_{(i, k-1)}\) is the average vector difference for the \(i\)-th category at the \(k\)-th iteration, calculated as:

\begin{equation}
\label{eq:proxy_delta}
\Delta P_{(i, k-1)} = \frac{1}{|F_i|} \sum_{\mathbf{f} \in F_i} (\mathbf{f} - P_{(i, k-1)})
\end{equation}

In Equation~\ref{eq:proxy_delta}, \(\Delta P_{(i, k-1)}\) represents the update applied to the class proxy for the \(i\)-th category at iteration \(k-1\). The set \(F_i\) consists of nodes within the current batch that are predicted by the CLIP model to belong to the \(i\)-th category. This prediction is based on the logits \(W\), as determined by Equation~\ref{eq:lt}, and applies to both teacher (\(P^t\)) and student (\(P^s\)) class proxies. The hyperparameter \(\alpha\) governs the update's magnitude, ensuring that the class proxies are adjusted appropriately.

For each category, the class proxy nodes of both teacher and student are updated by the rule defined in Eq.~\ref{eq:proxy_update}, where the average vector difference \(\Delta P_{(i, k-1)}\) is calculated as per Eq.~\ref{eq:proxy_delta}. This process ensures that the updates are specific to the category and reflective of the current batch's nodes embeddings.
The hyperparameter \(\alpha\) controls the update magnitude, which is chosen based on the dataset and batch size. Typically, \(\alpha\) is set approximately to the ratio of the batch size to the total number of training samples. This choice of \(\alpha\) ensures that the class proxies are updated to balance adaptability with stability. Ablation studies in Section~\ref{sec:alpha} comparing different settings of the hyperparameter \(\alpha\) have demonstrated the effectiveness of this approach. Specifically, the chosen value of \(\alpha\) allows the class proxy nodes to adapt to the evolving node embedding space while maintaining stability. This balance is crucial for accurately capturing the relationships between sample embeddings and class proxies, as it ensures that the PRG remains responsive to new information without being overly influenced by any single batch of data.

\noindent\textbf{Edge Formation.} We calculate edges based on the Pearson correlation coefficients (PCC) \cite{pearson1896mathematical} between node embeddings. For two vectors \(\mathbf{x}\) and \(\mathbf{y}\) in sets \(F\) and \(P\), respectively, the edge \(e_{xy}\) is computed as Equation~\ref{eq:pcc}:
\begin{equation}
\begin{aligned}
e_{x,y} = \frac{\sum_{i=1}^{D}(x_{i}-\bar{\mathbf{x}})(y_{i}-\bar{\mathbf{y}})}
{\sqrt{\sum_{i=1}^{D}(x_{i}-\bar{\mathbf{x}})^{2}} \sqrt{\sum_{i=1}^{D}(y_{i}-\bar{\mathbf{y}})^{2}}}
\end{aligned}
\label{eq:pcc}
\end{equation}
where \(x_{i}\) and \(y_{i}\) are the \(i\)-th elements of node embeddings \(\mathbf{x}\) and \(\mathbf{y}\) respectively, and \(\bar{\mathbf{x}}\) and \(\bar{\mathbf{y}}\) are their means. This formula facilitates the construction of edges based on the correlation between embeddings, capturing the relational knowledge between instances and classes within the graph.
Compared to calculating the correlation between instance nodes (\(F\)-\(F\)), we calculate the correlation between instances and class proxies (\(F\)-\(P\)) to be able to filter out the redundant knowledge of inter-instance correlation that is not relevant to the task at hand.
Thus, the PRG integrates instance nodes, class proxy nodes, and the edges between them, representing both the instance knowledge in sample features and logits, and the relational knowledge between batch samples and all categories within the target task. This comprehensive approach facilitates a deeper understanding of the structural dynamics within the embedding space, which is crucial for effective knowledge distillation. Further details on utilizing this structure for distillation will be discussed in subsequent sections.

\subsection{Alignment of the PRG}
\label{sec:alignment}
As discussed in Section~\ref{sec:graph}, we begin by leveraging the textual modality of CLIP to extract task-relevant knowledge using prompt-based weighted logits. This knowledge is then integrated into node embeddings within the integrated embedding space. Subsequently, we utilize edges between class proxy nodes and sample nodes to filter out task-irrelevant information. By doing so, we enhance the task-relevant knowledge from the CLIP teacher, thereby creating the Proxy Relational Graph (PRG). By aligning the student's PRG with that of the teacher, we enable the student network to absorb this complex structural information. The subsequent sections will detail the alignment of PRG nodes and edges,then describe our approach to formulating distillation objectives for aligning nodes and edges.

\begin{table*}[t]
  \caption{\textbf{Performance comparison} on other general-scale datasets. \textnormal{Abbreviations ``EN'', ``RV'', ``FV'' are used forEfficientNet-B1~\cite{tan2019efficientnet}, RepVit M1.1~\cite{ wang2023repvit}, and FastViT T12~\cite{vasu2023fastvit} models, respectively.}}
  \centering
  \resizebox{\textwidth}{!}{
    \begin{tabular}{lccccccccc}
    \toprule
          & \multicolumn{3}{c}{CIFAR10\cite{alex2009learning}} & \multicolumn{3}{c}{CIFAR100\cite{alex2009learning}} & \multicolumn{3}{c}{Caltech101\cite{fei2004learning}} \\
    \cmidrule(lr){2-4} \cmidrule(lr){5-7} \cmidrule(lr){8-10}
    Method & EN-B1\cite{tan2019efficientnet}
 & RV M1.1\cite{wang2023repvit}
 & FV T12\cite{vasu2023fastvit}
 & EN-B1\cite{tan2019efficientnet}
 & RV M1.1\cite{wang2023repvit}
 & FV T12\cite{vasu2023fastvit}
 & EN-B1\cite{tan2019efficientnet}
 & RV M1.1\cite{wang2023repvit}
 & FV T12\cite{vasu2023fastvit}
 \\
    \midrule
    CLIP ViT-L/14 \cite{radford2021learning} & \multicolumn{3}{c}{96.20} & \multicolumn{3}{c}{77.90} & \multicolumn{3}{c}{92.60} \\
    \midrule
    KD \cite{hinton2015distilling} & 94.43 & 94.99 & 94.81 & 74.61 & 74.81 & 72.05 & 81.16 & 81.05 & 81.11 \\
    DIST \cite{huang2022knowledge} & 94.46 & 95.20 & 94.51 & 74.49 & 72.65 & 71.75 & 80.47 & 80.07 & 77.88 \\
    FitNets \cite{adriana2015fitnets} & 94.88 & 95.33 & 95.16 & 73.89 & 73.43 & 73.06 & 80.13 & 77.65 & 79.61 \\
    AT \cite{zagoruyko2017paying}   & 95.57 & 95.56 & 94.80 & 75.31 & 73.17 & 73.15 & 79.49 & 79.44 & 79.49 \\
    RKD \cite{park2019relational}  & 92.93 & 94.74 & 94.40 & 72.05 & 72.25 & 72.10 & 80.47 & 79.90 & 79.72 \\
    \midrule
    \textbf{PRG(ours)} & \textbf{95.71} & \textbf{95.74} & \textbf{95.50} & \textbf{75.82} & \textbf{76.23} & \textbf{73.36} & \textbf{81.51} & \textbf{83.47} & \textbf{81.39} \\
    \bottomrule
    \end{tabular}%
    }
  \label{tab:normal}%
\end{table*}%

\noindent\textbf{Edge Alignment.}
In PRG, the edge matrix is constructed to represent the relationships between samples and class proxies. This matrix is formulated based on Eq.~\ref{eq:pcc}. Let \(e_{i,j}\) denote the correlation between any sample and class proxy nodes in the graph. The edge matrix for PRG can be expressed as:
\begin{equation}
\begin{aligned}
E(F, C) = \left(e_{i,j}\right) \in \mathbb{R}^{b \times c}
\end{aligned}
\label{eq:edgematrixPRG}
\end{equation}
where \(b\) is the batch size and \(c\) is the number of classes. The edge matrices for the teacher and student models, \(E^{t}\) and \(E^{s}\), are computed using their respective node sets, which include both instance and class proxy nodes. Thus, the constructed \(E^{t}\) calculates the relationships between each instance and all target task classes, focusing on the discriminative knowledge of each instance in the target task, effectively filtering out task-irrelevant knowledge.
To align the structural information captured in the PRG of the teacher and student models, we define an edge matching loss for PRG. This loss aims to align the edge matrices of the teacher and student graphs, as defined below:
\begin{equation}
\begin{aligned}
\mathcal{L}_{edge} =
{||{E}^{t} - {E}^{s}||_2}
\end{aligned}
\label{eq:edgematchPRG}
\end{equation}
This loss encourages the student network to mimic the structural correlations, including those involving class proxies, learned by the teacher network.

\noindent\textbf{Node Alignment.} In PRG, the alignment of sample node embeddings between the teacher and student networks is achieved through a correlation matrix, denoted as \(E(F^t, F^s)\).This matrix, computed using the Pearson Correlation Coefficient similar to Equation~\ref{eq:edgematrixPRG}, captures the cross-correlations between corresponding sample nodes from both networks. Notably, the class proxy node, which plays a crucial role in extracting relational knowledge, is not the primary focus of node alignment and is consequently not designed to participate in backpropagation processes. The node matching loss is defined as:
\begin{equation}
\mathcal{L}_{node} = {||{E(F^t, F^s) - \mathcal{I}}||_{2}}
\end{equation}
This loss function aims to maximize the correlation between corresponding nodes across the teacher and student networks, ensuring that the student model effectively mimics the teacher model's representations at the node level.

\noindent\textbf{Full Graph Alignment.}
Finally, to align the PRGs of the teacher and student models, we combine the edge and node matching losses into a single loss function for PRG distillation:
\begin{equation}
\begin{aligned}
\mathcal{L}_{PRG} = \lambda_{node}\mathcal{L}_{node} +\lambda_{edge} \mathcal{L}_{edge}
\end{aligned}
\label{eq:gmPRG}
\end{equation}
Where \(\lambda_{node}\) and \(\lambda_{edge}\) are hyperparameters used to adjust the contributions of the two loss terms respectively, this combined loss function ensures that the structural relationships and sample node embeddings are effectively aligned between the teacher and student networks within the PRG framework.

\subsection{Distilling students via Proxy Relational Graph}
\label{sec:distill}

\noindent\textbf{Overall Objective.}
The primary objective in our approach involves a standard cross-entropy loss, \(\mathcal{L}_{CE}\), using the softmax-transformed CLIP output logits acting like a conventional supervisory signal, and a PRG alignment loss, \(\mathcal{L}_{PRG}\), for effective task-relevant knowledge distillation. The overall objective of our model can be expressed as:

\begin{equation}
\begin{aligned}
\mathcal{L} = \mathcal{L}_{CE} + \mathcal{L}_{PRG} =\mathcal{L}_{CE} +\lambda_{node}\mathcal{L}_{node} +\lambda_{edge} \mathcal{L}_{edge}
\end{aligned}
\label{eq:crg_final}
\end{equation}

\section{Experiments}
\subsection{Experimental Settings}

\noindent\textbf{Datasets and setup}. We assess our framework and PRG method in annotation-free image classification tasks. To facilitate comparison and analysis of the method's effectiveness, we conduct experiments on traditional supervised datasets. Throughout the entire experimental process, we adhere to the following principle: during the distillation training phase, we only use sample images from the training set, without leveraging their original annotations in any manner. We solely test the accuracy of the student models on the validation and test sets. We primarily select three types of classification datasets for evaluation. The first type includes fine-grained classification datasets, such as Stanford Cars\cite{krause20133d}, Oxford Pets\cite{parkhi2012cats}, and Food-101\cite{bossard2014food}. The second type is represented by the large-scale dataset ImageNet-1K\cite{deng2009imagenet}. The third type encompasses datasets of a general scale, including CIFAR-10\cite{alex2009learning}, CIFAR-100\cite{alex2009learning}, and Caltech-101\cite{fei2004learning}. The choice of these datasets is mainly to investigate the validity of our motivation, as we hypothesize that excluding task-irrelevant knowledge from the LFM teacher's output should theoretically improve the model's performance, particularly in the context of fine-grained classification. Additionally, we aim to explore the effectiveness of our method in general-scale datasets and larger datasets such as ImageNet-1K\cite{deng2009imagenet}, to ascertain its applicability across various classification scenarios.

\noindent\textbf{Baseline}
We compare our method with three kinds of methods, namely the logit-based method KD~\cite{hinton2015distilling}, the relation-based method DIST~\cite{huang2022knowledge}, and the feature-based methods FitNet~\cite{adriana2015fitnets}, AT~\cite{zagoruyko2017paying}, and RKD~\cite{park2019relational}.

\noindent\textbf{Teacher and Student Models.} We adopt a ViT-L/14 model from CLIP \cite{radford2021learning} as the teacher network. For the student models, we select three efficient yet compact models: EfficientNet-B1~\cite{tan2019efficientnet}, RepVit M1.1~\cite{ wang2023repvit}, and FastViT T12~\cite{vasu2023fastvit}.

\noindent\textbf{Implementation Details.}
The standard RandAugment technique \cite{cubuk2020randaugment} is applied for data augmentation. Key hyperparameters in our experiments, including \(\lambda_{edge}\) set to 0.2 and \(\lambda_{node}\) set to 0.4, are optimized through ablation studies conducted on the CIFAR-100\cite{alex2009learning} dataset. The training process utilizes the AdamW optimizer with an initial learning rate of 0.03, paired with a cosine scheduler \cite{loshchilov2016sgdr} having \(T_0=10\) and \(T_{\text{mult}}=2\), spanning a total of 150 epochs with a batch size of 64. For experiments conducted on ImageNet-1K\cite{deng2009imagenet}, the initial learning rate is adjusted to 0.001, and the scheduler settings are modified to \(T_0=8\) and \(T_{\text{mult}}=2\), over a training duration of 120 epochs. For all datasets except OxfordPets and Food101, we used the official prompts provided in the CLIP paper, aligning with the original settings. For OxfordPets and Food101, we added CIFAR100 prompts to maintain the integrity of the weighted logits. Unlike standard CLIP averaging, our method applies a weighted average to focus more on prompts with higher confidence.

\begin{table*}[htbp]
  \caption{\textbf{Performance comparison} on the fine-grained datasets. \textnormal{Abbreviations ``EN'', ``RV'', ``FV'' are used for EfficientNet-B1~\cite{tan2019efficientnet}, RepVit M1.1~\cite{ wang2023repvit}, and FastViT T12~\cite{vasu2023fastvit} models, respectively.}}
  \centering
    \resizebox{\textwidth}{!}{
    \begin{tabular}{lccccccccc}
    \toprule
          & \multicolumn{3}{c}{StanfordCars\cite{krause20133d}} & \multicolumn{3}{c}{OxfordPets\cite{parkhi2012cats}} & \multicolumn{3}{c}{Food101\cite{bossard2014food}}\\
    \cmidrule(lr){2-4} \cmidrule(lr){5-7} \cmidrule(lr){8-10}
    Method & EN-B1\cite{tan2019efficientnet} & RV M1.1\cite{wang2023repvit} & FV T12\cite{vasu2023fastvit} & EN-B1\cite{tan2019efficientnet} & RV M1.1\cite{wang2023repvit}
 & FV T12\cite{vasu2023fastvit} & EN-B1\cite{tan2019efficientnet} & RV M1.1\cite{wang2023repvit} & FV T12\cite{vasu2023fastvit} \\    
    \midrule
    CLIP ViT-L/14 \cite{radford2021learning} & \multicolumn{3}{c}{77.30} & \multicolumn{3}{c}{93.50} & \multicolumn{3}{c}{92.90} \\
    \midrule
    KD \cite{hinton2015distilling}   & 57.75 & 73.52 & 73.36 & 80.92 & 83.35 & 78.39 & 84.08 & 87.47 & 85.16 \\
    DIST \cite{huang2022knowledge} & 73.50 & 73.70 & 72.77 & 82.31 & 80.10 & 77.95 & 84.50 & 85.28 & 84.53 \\
    FitNets \cite{adriana2015fitnets} & 68.35 & 72.12 & 72.59 & 83.62 & 82.07 & 79.01 & 84.76 & 87.02 & 85.03 \\
    AT \cite{zagoruyko2017paying}   & 70.43 & 73.30 & 71.93 & 75.55 & 76.78 & 76.45 & 86.37 & 86.50 & 84.62 \\
    RKD \cite{park2019relational}  & 62.70 & 72.45 & 71.07 & 67.33 & 80.27 & 75.85 & 83.42 & 86.76 & 84.28 \\
    \midrule
    \textbf{PRG(ours)} & \textbf{77.32} & \textbf{74.13} & \textbf{73.59} & \textbf{84.60} & \textbf{84.22} & \textbf{79.72} & \textbf{87.03} & \textbf{87.71} & \textbf{85.31} \\
    \bottomrule
    \end{tabular}%
    }
  \label{tab:finegrain}%
\end{table*}%

\subsection{Results and Discussion.}
\subsubsection{Experiments on the general-scale dataset}

To validate the effectiveness of our method for annotation-free distillation in general classification domains using general-scale datasets, we conduct comparative experiments on CIFAR10\cite{alex2009learning}, CIFAR100\cite{alex2009learning}, and Caltech101\cite{fei2004learning}, as illustrated in Table~\ref{tab:normal}. In these comparisons, we observe the following: Our PRG method achieves state-of-the-art performance. Particularly on the CIFAR10\cite{alex2009learning} dataset, our method achieves an accuracy 95.74\%, only 0.46\% lower than that of the CLIP teacher. On the Caltech-101\cite{fei2004learning} dataset, despite its smaller trainset size compared to the CIFAR\cite{alex2009learning} datasets, our method still outperforms other distillation techniques. However, the performance gap between the student models and the CLIP teacher is not as narrow as on the CIFAR\cite{alex2009learning} datasets.

Based on these observations, our approach enables the student models to effectively assimilate task-specific discriminative knowledge from the CLIP teacher under annotation-free conditions, thus narrowing the performance gap with the CLIP teacher.

\subsubsection{Experiments on the fine-grained dataset}
To validate the challenges of task-irrelevant knowledge in the annotation-free distillation of LFMs, experiments were conducted on fine-grained classification datasets including StanfordCars\cite{krause20133d}, OxfordPets\cite{parkhi2012cats}, and Food101\cite{bossard2014food}, as shown in Table~\ref{tab:finegrain}. The observations from these experiments indicate the following:
First, our PRG method demonstrated a distinct advantage over all compared methods across all student models.
Second, a significant reduction in convergence speed was noted for some comparative KD methods on the StanfordCars\cite{krause20133d} and OxfordPets\cite{parkhi2012cats} datasets when using EfficientNet-B1~\cite{tan2019efficientnet} as the student model, resulting in lower performance under the same experimental setup. In contrast, our PRG method achieved distillation results on the StanfordCars\cite{krause20133d} dataset that even surpassed the zero-shot classification accuracy of the CLIP teacher.
Third, in a large part of the settings, particularly with methods that use only features or attention based on features, like FitNet~\cite{adriana2015fitnets} and AT~\cite{zagoruyko2017paying} a performance decline is observed compared to the logit-based KD method. Specifically, the results from AT~\cite{zagoruyko2017paying}, an attention-based distillation method, indicate that directly modeling the attention of the LFM teacher does not effectively extract the knowledge of the LFM for specific target tasks. This is because the attention weights include the LFM's attention to objects in general tasks, not only in target tasks.

These observations substantiate the severe challenge posed by task-irrelevant knowledge, as highlighted by the second point. The third point confirms the existence of the challenge posed by the high feature density of LFMs. Overall, the first observation validates that our method effectively addresses the two challenges identified. Additionally, the significant performance drop of comparison methods on fine-grained datasets compared to general datasets corroborates the negative impact of task-irrelevant information in LFM on specific tasks.

\begin{table}[hb]
    \caption{Performance comparison on the ImageNet-1K\cite{deng2009imagenet} dataset using RepViT M1.1 \cite{wang2023repvit} as the student model.}
  \centering
    \begin{tabular}{lc}
    \toprule
    Method & Accuracy (\%) \\
    \midrule
    CLIP ViT-L/14 \cite{radford2021learning} & 75.3 \\
    \midrule
    KD \cite{hinton2015distilling} & 71.15 \\
    DIST \cite{huang2022knowledge} & 71.62 \\
    FitNets \cite{adriana2015fitnets} & 71.29 \\
    AT \cite{zagoruyko2017paying} & 71.89 \\
    RKD \cite{park2019relational} & 71.61 \\
    \textbf{Ours} & \textbf{72.44} \\
    \bottomrule
    \end{tabular}%
  \label{tab:imagenet1k}%
\end{table}%

\subsubsection{Experiments on the large-scale dataset}
To validate the effectiveness of our method on larger datasets, we conduct experiments on the ImageNet-1K\cite{deng2009imagenet} dataset. The results, as shown in Table~\ref{tab:imagenet1k}, demonstrate that our method surpasses other approaches in an annotation-free setting. It is crucial to note that despite the extensive scope of a dataset like ImageNet-1K\cite{deng2009imagenet}, which might seem to dilute the impact of task-irrelevant knowledge, the 1000 categories within ImageNet still represent a distinct task domain. This becomes particularly evident when comparing it to the semi-supervised, larger datasets used to train the CLIP model. Our method's superior performance on ImageNet-1K\cite{deng2009imagenet} highlights its efficacy in distilling and utilizing task-relevant knowledge efficiently on larger scale datasets.

\subsection{Ablation study.}

We conduct a comprehensive set of ablation studies to investigate the impact of each component within the PRG framework. These experiments are performed on the CIFAR-100\cite{alex2009learning} dataset usingRepViT M1.1 \cite{wang2023repvit} and EfficientNet-B1 \cite{tan2019efficientnet} as the student models and CLIP ViT-L/14~\cite{radford2021learning}  as the teacher. 

\begin{figure}[htbp]
  \centering
   \includegraphics[width=1\linewidth]{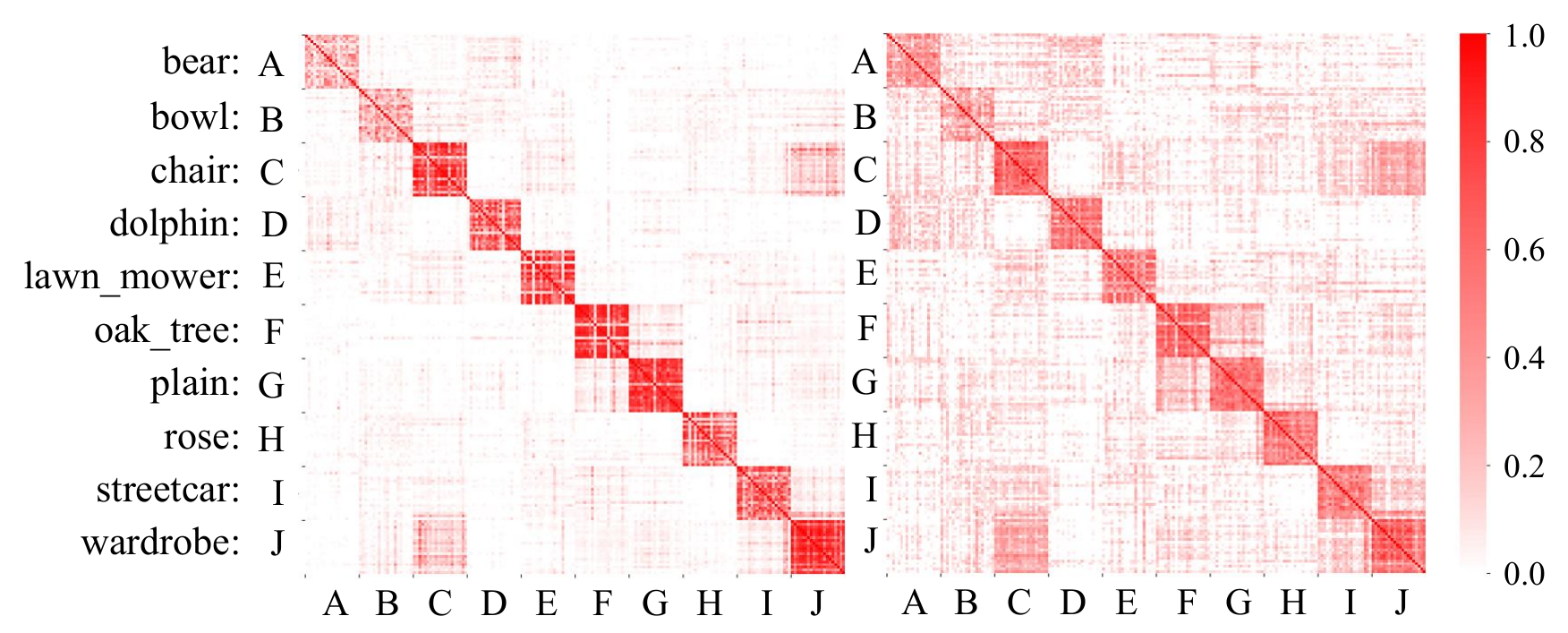}
   
    \caption{Heatmaps of inter-sample correlation on CIFAR100\cite{alex2009learning}, featuring the student model RepViT M1.1 \cite{wang2023repvit} (\emph{left}) and the teacher model ViT-L/14 from CLIP \cite{radford2021learning} (\emph{right}).}
   \label{fig:correlation_feat}
\end{figure}

\subsubsection{Task-irrelevant knowledge filtering}
To demonstrate how our method filters task-irrelevant knowledge from CLIP, we calculate the correlation coefficients of feature embeddings between teacher and student models across different samples, especially from different categories. We randomly select 20 samples from 10 categories within the CIFAR100\cite{alex2009learning} dataset and compute the Pearson correlation coefficients among these samples, as shown in Fig.~\ref{fig:correlation_feat}. The figure clearly illustrates the differences between the student model generated by PRG and the CLIP teacher. Notably, the correlation coefficients between feature embeddings produced by the student model are significantly lower than those generated by the CLIP teacher for different category samples. Since correlation coefficients indicate the similarity between features, lower similarity between features of different categories facilitates correct classification for specific tasks. Thus, based on these observations, we believe that the PRG method effectively filters out task-irrelevant inter-sample relational knowledge from CLIP, successfully addressing the challenge.

\begin{figure}[h]
\centering
\includegraphics[width=0.45\textwidth]{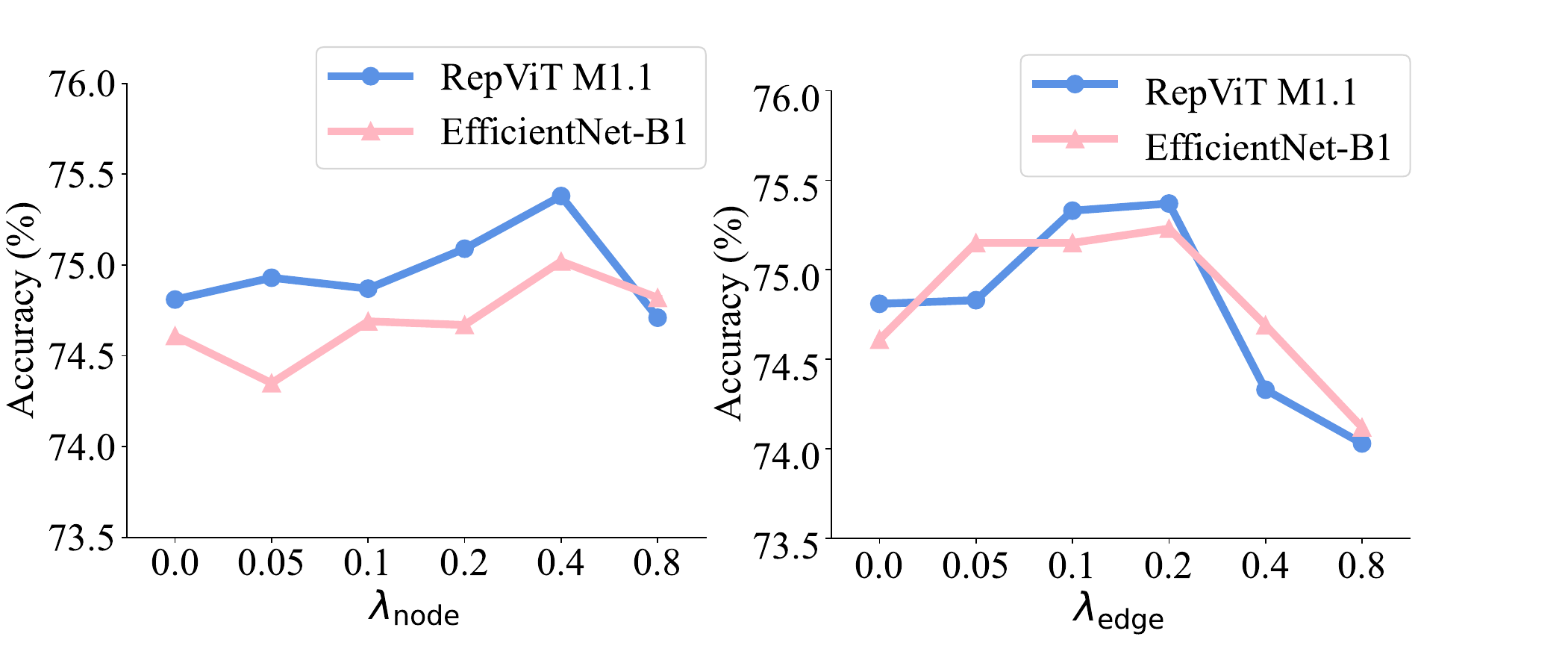}
\caption{Ablation study evaluating the effectiveness of node and edge alignment at different \(\lambda_{\text{node}}\) and \(\lambda_{\text{edge}}\) settings on CIFAR100\cite{alex2009learning}. The experiments involve RepViT M1.1 \cite{wang2023repvit} and EfficientNet-B1 \cite{tan2019efficientnet} as student models.}
\label{fig:abs1}
\end{figure}

\subsubsection{The impact of node and edge alignment}
To verify the effectiveness of node alignment and edge alignment within our PRG method, we conduct experiments by setting the \(\lambda\) value of one component to zero and adjusting the \(\lambda\) of the other. The results, illustrated in Figure~\ref{fig:abs1}, lead to two key observations. Firstly, in terms of node alignment, we find that as \(\lambda_{\text{node}}\) increases, the accuracy first improves and then decreases, with both student models exhibiting the highest accuracy at \(\lambda_{\text{node}} = 0.4\). 
Secondly, in our edge alignment experiments, we observe a similar trend in accuracy with changes in \(\lambda_{\text{edge}}\) as with node alignment. The optimal performance is achieved at \(\lambda_{\text{edge}} = 0.2\). Consequently, in our experiments, we set the weights for node alignment loss and edge alignment loss at 0.4 and 0.2, respectively.

\begin{table}[htbp]
  \caption{Impact of varying the update intensity coefficient (\(\alpha\)) for integrated sample nodes towards class proxy nodes in PRG, conducted on CIFAR-100\cite{alex2009learning} with a batch size of 64.}
  \centering
    \begin{tabular}{ccc}
    \toprule
    \(\alpha\) & RepViT M1.1 \cite{wang2023repvit} (\%) & EfficientNet-B1 \cite{tan2019efficientnet} (\%) \\
    \midrule
    1e-4 & 75.90 & 75.77 \\
    \textbf{1e-3} & \textbf{76.23} & \textbf{75.82} \\
    1e-2  & 73.54 & 72.66 \\
    \bottomrule
    \end{tabular}%
  \label{tab:alpha_update_intensity}%
\end{table}%

\subsubsection{Impact of update intensity for class proxy nodes}
\label{sec:alpha}
We aim to validate the setting of the update intensity \(\alpha\) for class proxy nodes, as detailed in Equation~\ref{eq:proxy_update}. To achieve this, we conduct experiments on CIFAR100\cite{alex2009learning} using three different \(\alpha\) values. The results are presented in Table~\ref{tab:alpha_update_intensity}. We observe that with a batch size of 64, the optimal performance of PRG is achieved when \(\alpha = 1 \times 10^{-3}\). According to our methodological design outlined in Section~\ref{sec:graph}, \(\alpha\) is expected to be \(1.28 \times 10^{-3}\) under these conditions, which is very close to the experimental result. This observation preliminarily confirms that our method for determining the value of \(\alpha\) can effectively and stably update the class proxy nodes.

\begin{table}[htbp]
  \caption{Ablation study on PRG method performance using feature embeddings(FE) vs. integrated embeddings(IE) on CIFAR-100\cite{alex2009learning}.}
  \centering
    \begin{tabular}{lcc}
    \toprule
    Configuration & RepViT M1.1 \cite{wang2023repvit} (\%) & EfficientNet-B1 \cite{tan2019efficientnet} (\%) \\
    \midrule
    PRG \(w\) FE & 75.53 & 74.76 \\
    \textbf{PRG \(w\) IE} & \textbf{76.23} & \textbf{75.82} \\
    \bottomrule
    \end{tabular}%
  \label{tab:embedding_comparison}%
\end{table}%

\subsubsection{Effectiveness of intergrated nodes}
To evaluate the effectiveness of using integrated embeddings as sample nodes, we conduct a comparative experiment using only feature embeddings as sample nodes. The experimental results, as depicted in Table~\ref{tab:embedding_comparison}, demonstrate that the PRG method performs better when integrated embeddings are used. This confirms the effectiveness of integrated embeddings in enhancing the knowledge distillation process.

\begin{table}[htbp]
  \caption{Ablation study on CIFAR-100\cite{alex2009learning} to evaluate the impact of using weighted logits (\(W\)) based on prompts in the PRG method. WL denotes weighted logits.}
  \centering
    \begin{tabular}{lcc}
    \toprule
    Configuration & RepViT M1.1 \cite{wang2023repvit} (\%) & EfficientNet-B1 \cite{tan2019efficientnet} (\%) \\
    \midrule
    PRG \(w/o\) WL & 75.73 & 75.33 \\
    \textbf{PRG \(w\) WL} & \textbf{76.23} & \textbf{75.82} \\
    \bottomrule
    \end{tabular}%
  \label{tab:cifar100_weighted_logits}%
\end{table}%

\subsubsection{Effectiveness of weighted logits}
To verify the effectiveness of our design, which employs weighted logits based on text embeddings for constructing teacher nodes, we conducted an experiment using the original CLIP logits to build the teacher's integrated sample nodes. The results, presented in Table~\ref{tab:cifar100_weighted_logits}, demonstrate that PRG with weighted logits performs better. This suggests that the design of weighted logits can effectively extract more task-relevant knowledge through the textual modality of the CLIP teacher, effectively addressing the challenge.

\section{Conclusion}
In this work, we present a novel framework for the annotation-free distillation of CLIP-like large foundational models (LFMs) across varied image classification tasks. We highlight the crucial challenge of filtering out task-irrelevant knowledge from LFMs during distillation. Our proposed solution, the Proxy Relational Graph (PRG) method, effectively addresses this by leveraging both the sample-category relationships within task domains and the task-specific semantic insights from CLIP's text prompt embeddings. This strategy enables PRG to overcome the constraints inherent in conventional distillation approaches.
Extensive evaluations across multiple datasets confirm PRG's superiority in closely mirroring the CLIP teacher's zero-shot capabilities, showcasing its effectiveness across general, large-scale, and fine-grained classification tasks. Remarkably, PRG achieves these results through an annotation-free process, introducing a cost-effective method for utilizing LFMs' vast knowledge, thereby facilitating the creation of efficient, lightweight models for diverse applications.
This research underscores the synergy between LFMs and distillation in an annotation-free context, encouraging further advancements in knowledge distillation. We anticipate that our findings will spur ongoing efforts to refine model training and deployment strategies, enhancing their efficiency and applicability.

\bibliographystyle{IEEEtran}
\bibliography{main}

\begin{IEEEbiography}[{\includegraphics[width=1in,height=1.25in,clip,keepaspectratio]{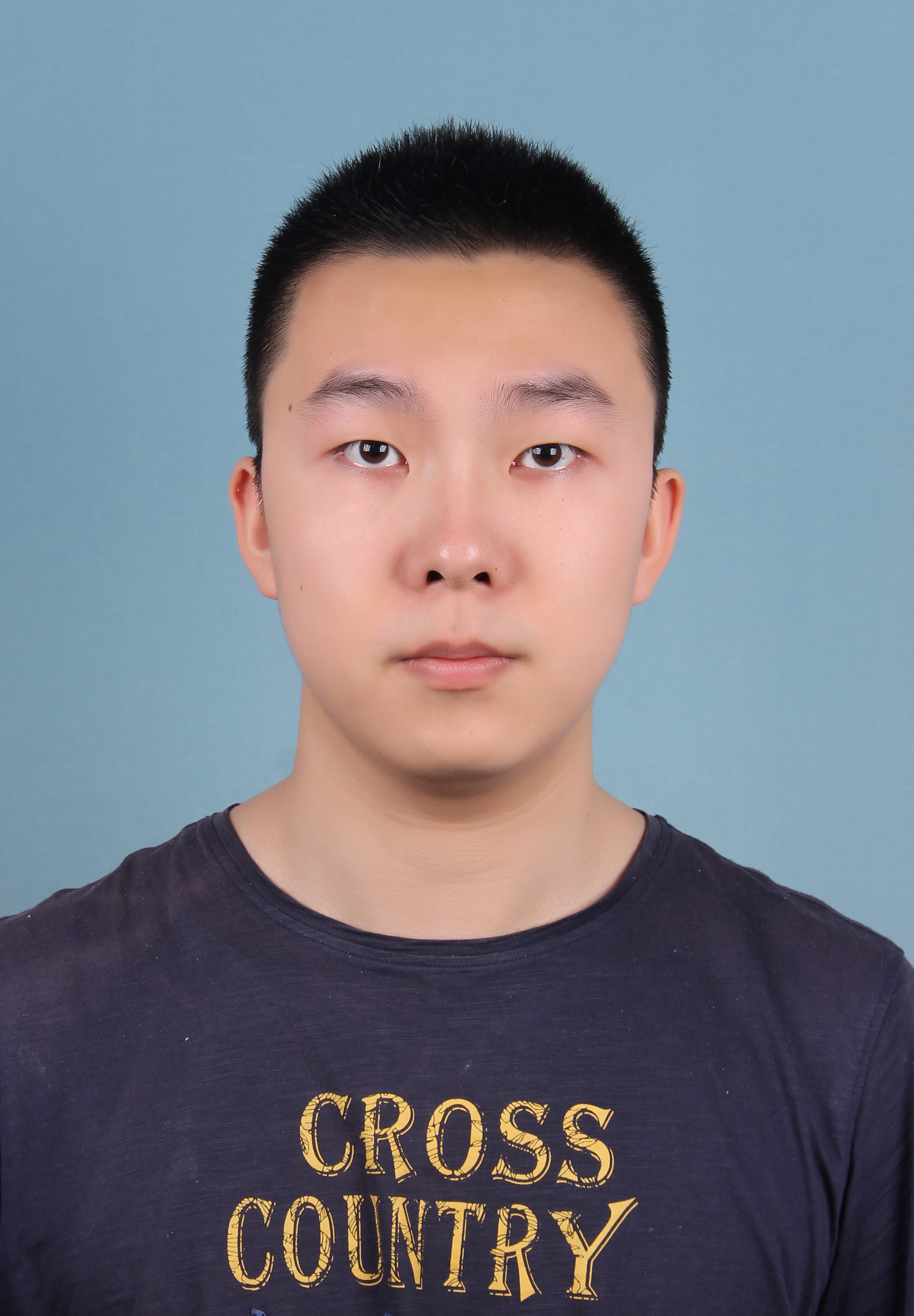}}]{Yijin Xu}
 is currently taking successive postgraduate and doctoral programs of study for Ph.D. degree. His current research interests are in knowledge distillation and transfer learning.

\end{IEEEbiography}

\begin{IEEEbiography}[{\includegraphics[width=1in,height=1.25in,clip,keepaspectratio]{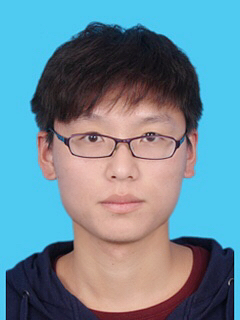}}]{Jialun Liu} received his Ph.D. degree from Jilin University and is currently continuing his research work at Baidu Vis. His research interest includes 3D generation.
\end{IEEEbiography}

\begin{IEEEbiography}[{\includegraphics[width=1in,height=1.25in,clip,keepaspectratio]{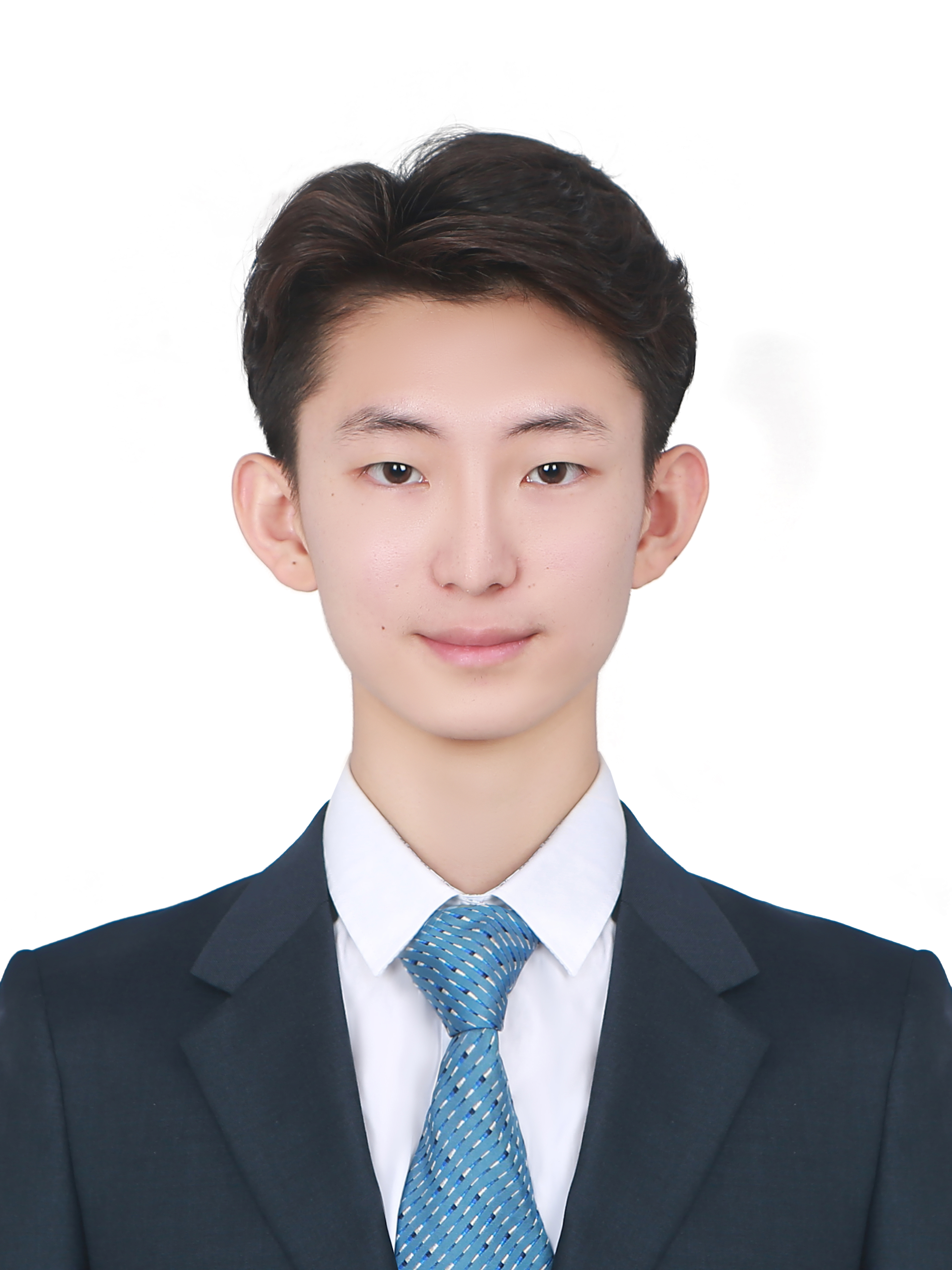}}]{Hualiang Wei} 
received the B.S. degree in Yanbian University, and is currently taking successive postgraduate and doctoral programs of study for Ph.D. degree in Jilin University. His research interests include 2D Image Synthesis, Avatar Generation and Animation and Image and Video Generation.
\end{IEEEbiography}

\begin{IEEEbiography}[{\includegraphics[width=1in,height=1.25in,clip,keepaspectratio]{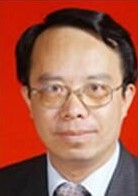}}]{Wenhui Li} 
is currently a Professor in the College of Computer Science and Technology, Jilin University. His research interests include pattern recognition and computer vision.
\end{IEEEbiography}


\vfill

\end{document}